\documentclass{article}

\usepackage{arxiv}

\usepackage[utf8]{inputenc} 
\usepackage[T1]{fontenc}    
\usepackage{hyperref}       
\usepackage{url}            
\usepackage{booktabs}       
\usepackage{amsfonts}       
\usepackage{nicefrac}       
\usepackage{microtype}      
\usepackage{graphicx}
\usepackage{natbib}
\usepackage{doi}

\usepackage{times}
\usepackage{soul}
\usepackage[small]{caption}
\usepackage{amsmath}
\usepackage{wrapfig}
\usepackage{float}

\usepackage{xcolor}
\usepackage{amssymb}

\title{Target Layer Regularization for Continual Learning Using Cramer-Wold Generator}

\author{{Marcin Mazur}\thanks{Corresponding author.} \\
	Faculty of Mathematics and Computer Science \\
	Jagiellonian University in Krakow \\
	Łojasiewicza 6, 30-348 Kraków, Poland\\
	\texttt{marcin.mazur@uj.edu.pl} \\
	\And
{Łukasz Pustelnik} \\
	Faculty of Mathematics and Computer Science \\
	Jagiellonian University in Krakow \\
	Łojasiewicza 6, 30-348 Kraków, Poland\\
	\texttt{lukas.pustelnik@student.uj.edu.pl} \\
	\And
{Szymon Knop} \\
	Faculty of Mathematics and Computer Science \\
	Jagiellonian University in Krakow \\
	Łojasiewicza 6, 30-348 Kraków, Poland\\
	\texttt{szymon.knop@doctoral.uj.edu.pl} \\
	\And
{Patryk Pagacz} \\
	Faculty of Mathematics and Computer Science \\
	Jagiellonian University in Krakow \\
	Łojasiewicza 6, 30-348 Kraków, Poland\\
	\texttt{patryk.pagacz@uj.edu.pl} \\
	\And
{Przemysław Spurek} \\
	Faculty of Mathematics and Computer Science \\
	Jagiellonian University in Krakow \\
	Łojasiewicza 6, 30-348 Kraków, Poland\\
	\texttt{przemyslaw.spurek@uj.edu.pl} \\
}

\date{}

\renewcommand{\headeright}{}
\renewcommand{\undertitle}{}

\hypersetup{
pdftitle={Target Layer Regularization for Continual Learning Using Cramer-Wold Generator},
pdfauthor={Marcin Mazur, Łukasz Pustelnik, Szymon Knop, Patryk Pagacz, Przemysław Spurek},
}

\def\our{CW-TaLaR}

\begin{document}

\maketitle

\begin{abstract}
We propose an effective regularization strategy (\our{}) for solving continual learning problems. It uses a penalizing term expressed by the Cramer-Wold distance between two probability distributions defined on a target layer of an underlying neural network that is shared by all tasks, and the simple architecture of the Cramer-Wold generator for modeling
output data representation.
Our strategy preserves target layer
distribution while learning a new task but does not require remembering previous tasks' datasets. We perform experiments involving several common supervised frameworks, which prove the competitiveness of the \our{} method in comparison to a few existing state-of-the-art continual learning models.

\end{abstract}

\section{Introduction}

The concept of {\em continual learning (CL)}, which aims to reduce the distance between human and artificial intelligence, seems to be considered recently by deep learning community as one of the main challenges. Generally speaking, it means the ability of the neural network to effectively learn consecutive tasks (in either supervised or unsupervised scenarios) while trying to prevent forgetting already learned information. Therefore, when designing an appropriate strategy, it needs to be ensured that the network weights are updated in such a way that they correspond to both the current and all previous tasks. However, in practice, it is quite likely that constructed CL model will suffer from either {\em intransigence} (hard acquiring new knowledge, see \cite{Chaudhry_2018_ECCV}) or {\em catastrophic forgetting (CF)} phenomenon (tendency to lose past knowledge, see \cite{MCCLOSKEY1989109}).

In recent years, methods of overcoming  the above-mentioned problems are subject to wide and intensive investigation. Following \cite{MALTONI201956}, the most popular CL strategies may be assigned into the following fuzzy categories:
\begin{itemize}
    \item[(i)] {\em architectural strategies}, involving specific (eventually growing) architectures and/or weight freezing/pruning,
    \item[(ii)] {\em (pseudo) rehearsal strategies}, in which past information is remembered, preferably  (to avoid increasing memory consumption) exploiting a generative model, and then replayed in future training,
    \item[(iii)] {\em regularization strategies}, introducing a penalization term into the loss function, which promotes selective consolidation of past information or slows training on new tasks.
\end{itemize}

As respective pure examples, we can recall here: (i) Progressive Neural Network (PNN, see \cite{rusu2016progressive}), (ii) Experience Replay for Streaming Learning (ExStream, see \cite{hayes2019memory}), and (iii) Learning without Forgetting (LwF, see \cite{Li2016}).

For a more complete and detailed overview of the existing CL models and their classification according to the above distinction rules, we refer the reader to \cite{MALTONI201956}, especially to the very informative Venn diagram provided as Figure 1.

This paper aims to show that it is achievable to construct an effective regularization strategy with a penalizing term expressed by the Cramer-Wold distance (introduced by \cite{JMLR:v21:19-560}) between two probability distributions designed to represent current and past information, both defined on a target layer of a neural network that is shared by all models dedicated to solve individual tasks. To memorize the past knowledge, an additional simple generator architecture is learned to retrieve the network output data from Gaussian noise.
Following \cite{JMLR:v21:19-560} we call it the Cramer-Wold (or CW) generator. 
It is worth noting that such strategy preserves the network output distribution while learning a new task, but does not require remembering/replaying (usually high dimensional) previous tasks' datasets. A general description of our approach, involving theoretical (mathematical) background, the reader can find in Section \ref{model}.

We apply our method (see Section \ref{classification}), which we call henceforth {\em Cramer-Wold Target Layer Regularization (CW-TaLaR)} strategy, for different categories of supervised benchmark CL learning scenarios proposed by \cite{hsu2018re}, involving {\em Incremental Task Learning}, {\em Incremental Domain Learning} and {\em Incremental Class Learning} on Split MNIST, Split CIFAR-10 and Permuted MNIST datasets. The conducted experiments show (see Section \ref{experiments}) that the results of the CW-TaLaR method are comparable or better then those obtained by the other existing state-of-the-art regularization-based CL models, including (online) Elastic Weights Consolidation (EWC, see \cite{Kirkpatrick3521}), Synaptic Intelligence (SI, see \cite{Zenke2017}) and  Memory Aware Synapses (MAS, see \cite{Aljundi_2018_ECCV}).

\begin{figure}[!t]
    \centering
    \includegraphics[width=0.85\textwidth]{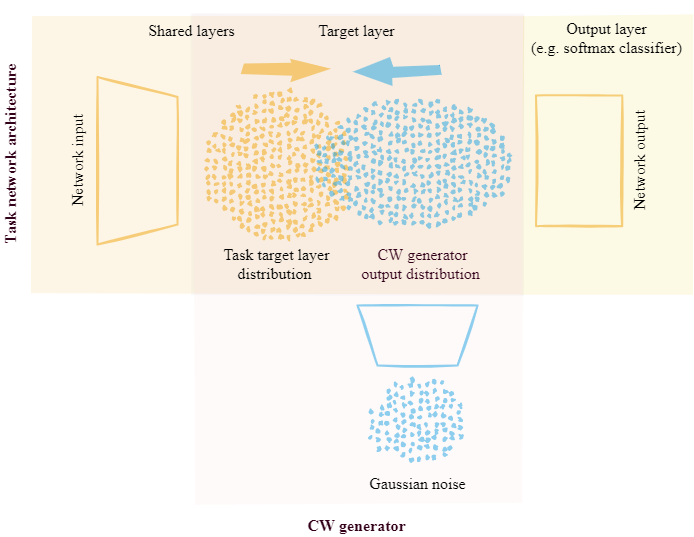}
    \caption{In each step of the proposed algorithm, we regularize our model by forcing closeness of task target layer distribution (representing current information) and CW generator output distribution (representing past information). This applies to both solving individual tasks as well as training the CW generator. Note that our construction does not require extra memory to remember previous tasks' datasets. (The figure was produced using  {\tt diagrams.net} software.)}
    \label{fig:model}
\end{figure}

\begin{figure}[!t]
\centering
\begin{tabular}{@{}c@{}c@{}c@{}}
First task &  Second task (no regularization)  & Second task (CW-TaLaR)  \\
\includegraphics[width=0.33\textwidth]{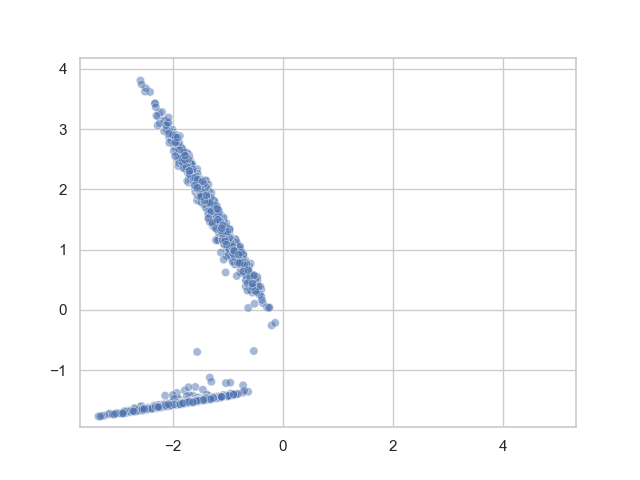} & \,
\includegraphics[width=0.33\textwidth]{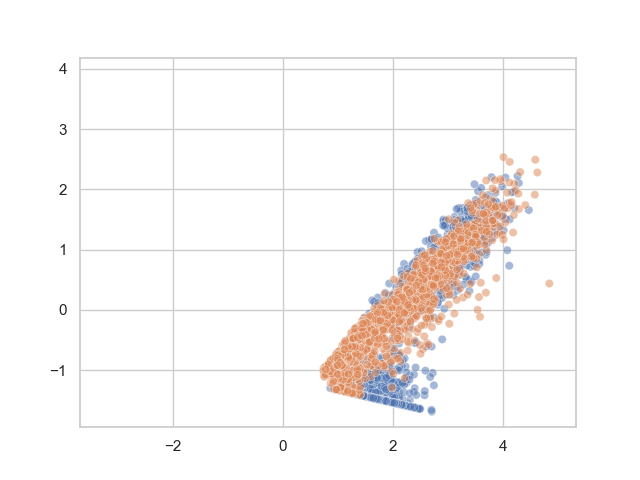} & \ 
\includegraphics[width=0.33\textwidth]{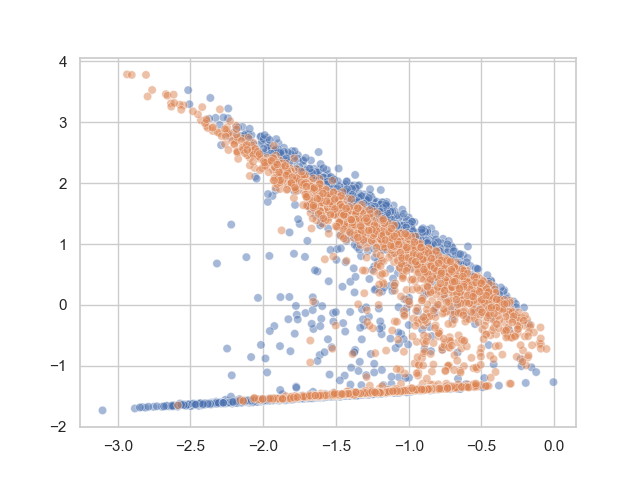} 
\end{tabular}
\caption{The figures demonstrate how CW-TaLaR works in the real world CL scenario. Blue dots represent output produced on a target layer for samples from the first task wile orange dots represent output for samples from the second task. In the first column, we present how output data look like after training the first task. In the second column, we show how output data behave when there is no regularization incorporated. Note that in such a scenario output for the first task samples completely changed their distribution. In the third column, we show how our method prevents output data from diverging from the previous tasks' distribution which eventually improves model continual learning capabilities. The above results were obtained in the Split MNIST experimental setting (see Section \ref{experiments}). Output data  dimensionality has been reduced with PCA for the purpose of illustration.    
}
\label{fig:intuition}
\end{figure}

A concept of our regularization strategy and an intuitive motivation (involving the real world CL scenario) for the \our{} model are presented in Figures \ref{fig:model} and \ref{fig:intuition}.

Our contribution can be summarized as follows:
\begin{itemize}
\item[(i)] we introduce a novel CL strategy (CW-TaLaR), which is based on the Cramer-Wold distance,
\item[(ii)] we compare the CW-TaLaR method with the other regularization-based models, including EWC, SI and MAS.
\end{itemize}

\section{Cramer-Wold distance and CW generator}\label{CW}

The key idea behind the Cramer-Wold distance ($d_\mathrm{cw}$) between two multi-dimensional distributions, which was firstly defined by \cite{JMLR:v21:19-560}, is to consider squared $L_2$ distance computed across multiple single-dimensional smoothed (with a Gaussian kernel) projections of their density functions. In a practical context, where we deal with two $D$-dimensional samples $X=(x_i)$ and $Y=(y_i)$, calculation of $d_\mathrm{cw}$ boils down to the following general formula:
\begin{equation}\label{dcw}
d^2_\mathrm{cw}(X,Y)=\int_{S^{D-1}}\|\mathrm{sm}_\gamma(v^TX)-\mathrm{sm}_\gamma(v^TY)\|^2_{2}\; \sigma_{D-1}(v),
\end{equation}
where ${\mathrm{sm}_\gamma}$ means a smoothing function using a Gaussian kernel with the variance $\gamma$, and $S^{D-1}$ denotes the unit sphere in $\mathrm{R}^D$ with the normalized surface measure $\sigma_{D-1}$. Then, in light of Theorem 1 from \cite{JMLR:v21:19-560} and assuming sufficiently large dimension of data ($D\geq 20$), in all numerical experiments we can use an approximate version of (\ref{dcw}), namely:
\begin{equation} 
\begin{array}{l @{} l}
2\sqrt{\pi}\,d^2_\mathrm{cw}(X,Y) \, \approx \;\; & \frac{1}{n^2}\sum_{ij}  (\gamma_n+\tfrac{\|x_i-x_j\|^2}{2D-3})^{-\frac{1}{2}} \,+ \, \tfrac{1}{n^2}\sum_{ij} 
(\gamma_n+\tfrac{\|y_i-y_j\|^2}{2D-3})^{-\frac{1}{2}} \\
                   &-\;\, \tfrac{2}{n} \sum_{ij} (\gamma_n+\tfrac{\|x_i-y_j\|^2}{2D-3})^{-\frac{1}{2}},
\end{array}
\end{equation}
where $\gamma_n$ is a bandwidth chosen using Silverman's rule of thumb\footnote{Specifically, $\gamma_n=\hat{\sigma}(\frac{4}{3n})^{2/5}$, where  $\hat{\sigma}$ denotes a~standard deviation calculated using joined $X$ and $Y$ samples.}.

Since the Cramer-Wold distance turned out to have a~closed-form for spherical Gaussians, it was used by \cite{JMLR:v21:19-560} to construct the Cramer-Wold autoencoder (CWAE). Moreover, the authors of CWAE indicated that it is also possible to apply $d^2_{\mathrm{cw}}(X,G(Z))$ as an objective function for a data generator $G$ (hence called the CW generator), mapping a latent Gaussian noise $Z$ into the data space. We want to emphasize that this simple concept showed promising results when trained on MNIST and Fashion MNIST datasets (see Section 8 of \cite{JMLR:v21:19-560}), which motivated us to consider it as a part of our model.

\section{Proposed algorithm}\label{model}

We deal with a general CL scenario, in which a sequence of tasks $T_1, \ldots, T_n$, using data samples $(X_1,Y_1)=(x^1_i,y^1_i), \ldots , (X_n,Y_n)=(x^n_i,y^n_i)$ drawn from some (unknown) distributions on data spaces $\mathcal{X}\times \mathcal{Y}_1, \ldots, \mathcal{X}\times \mathcal{Y}_n$, is given.
Let $f_\theta\colon \mathcal{X}\to \mathcal{S}\subset\mathbb{R}^D$ be a neural network shared by all architectures that we need to optimize in order to solve all the tasks one by one (i.e. a new task is collected only when the current task is over). Assume that a solution of each task $T_j$ for $j=1,\ldots, n$ may be obtained by minimizing an appropriate objective $\mathcal{I}_j(\theta)=\mathcal{L}_j(f_\theta(X_j),Y_j)$, where $\mathcal{L}_j\colon \mathcal{S}\times \mathcal{Y}_j\to \mathbb{R}$ means a respective loss.

In our model, we regularize each cost function $\mathcal{I}_j(\theta)$ for $j=2,\ldots, n$, by tending to add tuned squared Cramer-Wold distance calculated between two distributions for tasks $T_{j-1}$ and $T_j$, which are defined on the target layer $\mathcal{S}$ and represented by the output samples $f_{\theta^*_{j-1}}(X_{j-1})$ and $f_\theta(X_{j})$, where $\theta^*_{j-1}$ denotes the optimal parameter value achieved when $T_{j-1}$ is over. However, since we do not already have access to previous tasks' datasets when we solve task $T_j$, we render a missing sample $f_{\theta^*_{j-1}}(X_{j-1})$ using the CW generator network $G_\gamma\colon \mathcal{Z}\to \mathcal{S}$, which is trained independently to generate the output of $f_{\theta^*_{j-1}}$. More precisely, our general algorithm can be presented in the following steps.

\paragraph{Solving $T_1$.} We solve $T_1$ by minimizing $\mathcal{I}_1(\theta)$ and obtain $\theta^*_{1}=\mathrm{argmin}_\theta \mathcal{I}_1(\theta)$.

\paragraph{Training $G_\gamma$.} We train $G_\gamma$ by minimizing $d^2_{\mathrm{cw}}(f_{\theta^*_1}(X_1),G_\gamma(Z))$, where \mbox{$Z=(z_i)$} means a sample from $N(0,I)$ (a Gaussian noise on the latent $\mathcal{Z}$), and obtain $\gamma^*_{1}=\mathrm{argmin}_\gamma d^2_{\mathrm{cw}}(f_{\theta^*_1}(X_1),G_\gamma(Z))$.

\paragraph{Solving $T_2$.} We solve $T_2$ by minimizing 
\begin{equation}
\tilde{\mathcal{I}}_2(\theta)=\mathcal{I}_2(\theta)+\lambda d^2_{\mathrm{cw}}(G_{\gamma^*_1}(Z),f_\theta(X_{2})),
\end{equation}
where $\lambda$ is a hyperparameter\footnote{The hyperparameter $\lambda$ (which may be different in each step) aims to find a balance between intransigence and forgetting properties.}, and obtain $\theta^*_{2}=\mathrm{argmin}_\theta \tilde{\mathcal{I}}_2(\theta)$.

\paragraph{Training $G_\gamma$.} We retrain $G_\gamma$ by minimizing $d^2_{\mathrm{cw}}(f_{\theta^*_2}(X_2),G_\gamma(Z))$ and obtain $\gamma^*_{2}=\mathrm{argmin}_\gamma d^2_{\mathrm{cw}}(f_{\theta^*_2}(X_2),G_\gamma(Z))$.

\begin{center}
    {\bf \ldots}
\end{center}
\paragraph{Solving $T_j$.} We solve $T_j$ by minimizing \begin{equation}
\tilde{\mathcal{I}}_j(\theta)=\mathcal{I}_j(\theta)+\lambda d^2_{\mathrm{cw}}(G_{\gamma^*_{j-1}}(Z),f_\theta(X_{j}))
\end{equation}
 and obtain $\theta^*_{j}=\mathrm{argmin}_\theta \tilde{\mathcal{I}}_j(\theta)$.

\paragraph{Training $G_\gamma$.} We retrain $G_\gamma$ by minimizing $d^2_{\mathrm{cw}}(f_{\theta^*_j}(X_j),G_\gamma(Z))$ and obtain $\gamma^*_{j}=\mathrm{argmin}_\gamma d^2_{\mathrm{cw}}(f_{\theta^*_j}(X_j),G_\gamma(Z))$.

\begin{center}
    {\bf\ldots}
\end{center}
\paragraph{Solving  $T_n$.} We solve $T_n$ by minimizing \begin{equation}
\tilde{\mathcal{I}}_n(\theta)=\mathcal{I}_n(\theta)+\lambda d^2_{\mathrm{cw}}(G_{\gamma^*_{n-1}}(Z),f_\theta(X_{n}))
\end{equation}
and obtain $\theta^*_{n}=\mathrm{argmin}_\theta \tilde{\mathcal{I}}_n(\theta)$.

 Note that due to flexibility of the above construction and various ways of interpretation of its particular elements,
the proposed method is enough general to be applied to many typical supervised and unsupervised learning frameworks. However, since in further study we concentrate on application of our model in solving classification problems, in the next section we describe a more precise version of the proposed algorithm designed for a supervised setting.

 \section{Adjustment to classification problems}\label{classification}
Under notation from the previous section, assume that $T_1,\ldots, T_n$ is a sequence of classification tasks. In this case data space  $\mathcal{Y}_j$ is a collection of labels $l_1,\ldots, l_{m_j}$ (that may differ between tasks), and the last shared layer $\mathcal{S}$ (containing the output of $f_\theta$) is mapped into a logit layer $\mathbb{R}^{m_j}$ by some single layer network $\phi_\delta^j$. Moreover, each loss $\mathcal{L}_j$ is  expressed by the total cross-entropy evaluated over the label sample $Y=(y_i)$ relative to the output of a classifier (a softmax function) applied for a given logit sample $\phi_\delta^j(S)=\phi_\delta^j(s_i)$, i.e.:
\begin{equation}\label{loss}
\mathcal{L}_j(S,Y)=-\sum_i\sum_k \mathbb{I}_{\{y_i=l_k\}} \log \mathrm{softmax}(\phi_\delta^j(s_i)),
\end{equation}
where $\mathbb{I}_{\{\cdot\}}$ means a characteristic function.

The above approach covers all three categories of CL scenarios, i.e. Incremental Task Learning (ITL), Incremental Class Learning (ICL) and Incremental Domain Learning (IDL), which were introduced by \cite{hsu2018re} in relation to the differences between tasks, taking into account the marginal distributions of input data and target labels.
Hence, we can apply our method to many variants of experimental setup
that were  proposed in \cite{hsu2018re}. In the next section we describe considered experimental scenarios as well as present the obtained results, comparing them to those provided by the (online) EWC, SI and MAS models.

\section{Experiments}\label{experiments}

To evaluate effectiveness of the \our{} model in a CL setting, we adhere to experimental setups proposed by \cite{hsu2018re}. We compare our strategy with a selection of regularization-based methods, i.e. (online) EWC, SI and MAS, as well as with some vanilla baselines. We use (mentioned above) three classic scenarios: ITL, IDL and ICL, which are applied for a few datasets commonly used in various CL settings: Split MNIST, Split CIFAR-10 and Permuted MNIST.

In ITL the output layer is split into 5 separate heads that serve as binary classifiers (only one  active per task), each of which discriminates between 2 different classes.
On the other hand, ICL uses a single head that grows accordingly, from 2 to 10 every 2 outputs.
In turn, IDL consists of a single head that is shared by all tasks, which serve as a classifier that distinguishes between 2 (for Split MNIST and Split CIFAR-10) or 10 (for Permuted MNIST) different classes.

We repeated each experiment 10 times and reported the average accuracy over all tasks and runs for all applied strategies. (See the following three subsections for the detailed results.) It should be noted that for the CW-TaLaR method we added an additional penalizing term to the objective function (i.e. $L_2$ norm taken from network output sample on a target layer) that prevents large magnitudes which may occur due to potentially unlimited ReLU values.

\subsection{Split MNIST}
We summarize the results of all experiments performed on the Split MNIST dataset in Table \ref{split_mnit_table}. Note that \our{} significantly improves performance in single-head CL scenarios (IDL and ICL).

\begin{table}[!t]
\begin{center}
\caption{Average accuracy (\%, higher is better) with standard deviation achieved on validation dataset by each method  applied for Split MNIST in three CL scenarios, calculated over 5 tasks and 10 runs (at the end of the training process).}
\vskip2mm
\begin{tabular}{cccc}
\toprule
   Method & ITL &  IDL & ICL \\
\midrule
   Adam         & 93.50 $\pm$ 3.40 & 54.48 $\pm$ 0.80 & 19.88 $\pm$ 0.02 \\
   SGD          & 94.45 $\pm$ 2.24 & 75.83 $\pm$ 0.40 & 18.70 $\pm$ 0.19 \\
   Adagrad      & 97.06 $\pm$ 0.53 & 63.78 $\pm$ 3.38 & 19.55 $\pm$ 0.03 \\
   L2           & 95.39 $\pm$ 3.62 & 65.52 $\pm$ 2.89 & 19.44 $\pm$ 2.89 \\ 
\midrule
   EWC          & 94.18 $\pm$ 3.52 & 55.74 $\pm$ 1.25 & 19.83 $\pm$ 0.03  \\
   Online EWC   & 94.18 $\pm$ 4.98 & 58.66 $\pm$ 1.37 & 19.86 $\pm$ 0.02  \\
   SI           & 97.80 $\pm$ 1.43 & 65.66 $\pm$ 1.49 & 20.54 $\pm$ 0.78  \\
   MAS          & \bf 97.97 $\pm$ 1.15 & 72.24 $\pm$ 5.37 & 20.24 $\pm$ 0.95  \\ 
  CW-TaLaR (ours) & 97.53 $\pm$ 0.53 &\bf 80.64 $\pm$ 1.25 & \bf 38.65 $\pm$ 3.02 \\
\midrule
   Offline (upper bound) & 99.59 $\pm$ 0.09 & 98.69 $\pm$ 0.13 & 97.79 $\pm$ 0.20 
 \end{tabular}
\label{split_mnit_table}
\end{center}
\end{table}

\begin{figure}[!t]
\centering
\begin{tabular}{@{}c@{}c@{}c@{}}
ITL &  IDL  & ICL  \\[-1mm]
  \includegraphics[width=0.33\textwidth]{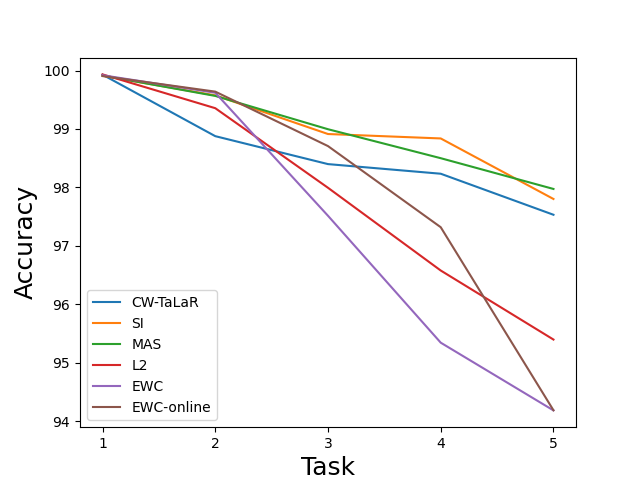}
& \
  \includegraphics[width=0.33\textwidth]{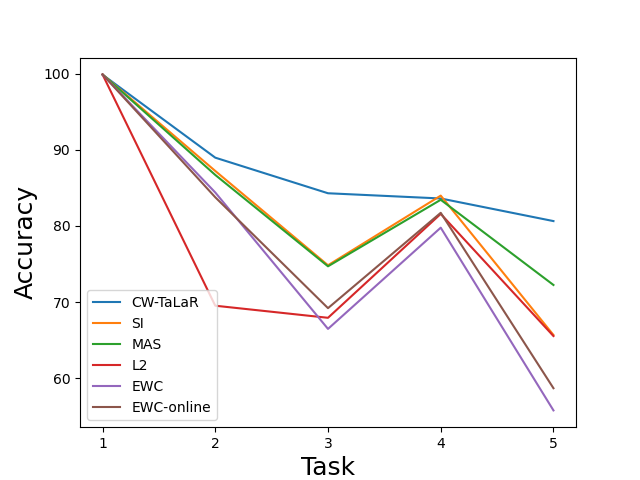}
& \
  \includegraphics[width=0.33\textwidth]{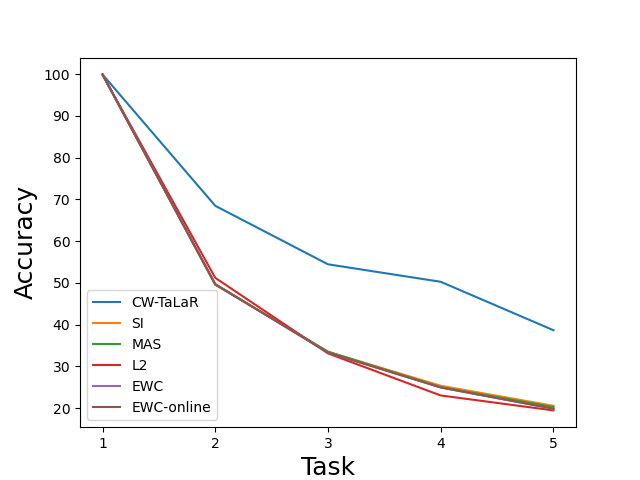}
\end{tabular}
\caption{Comparison of average accuracy over currently completed tasks achieved by each method applied for Split MNIST in three CL scenarios, calculated on validation dataset during the training process. (All results were additionally averaged over 10 runs.) Note that CW-TaLaR is superior to overcome forgetting in single-head scenarios.}
\label{fig:comp_forget_mnist}
\end{figure}

Figure \ref{fig:comp_forget_mnist} shows 
that our strategy effectively struggles with catastrophic forgetting, especially in the case of IDL setting, where all other methods fail to overcome this issue since their accuracy fall drastically  from time to time, as well as in the case of ICL setting, where significantly wins throughout the entire learning process.

\subsection{Split CIFAR-10}

\begin{figure}[!t]
\centering
\begin{tabular}{@{}c@{}c@{}c@{}}
ITL &  IDL  & ICL  \\[-1mm]
  \includegraphics[width=0.33\textwidth]{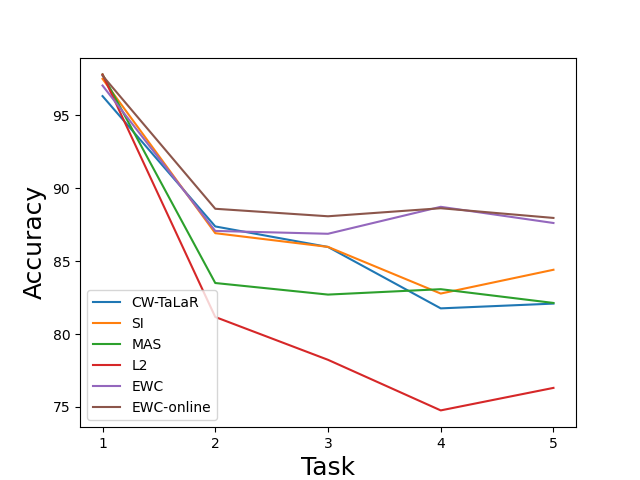}
& \
  \includegraphics[width=0.33\textwidth]{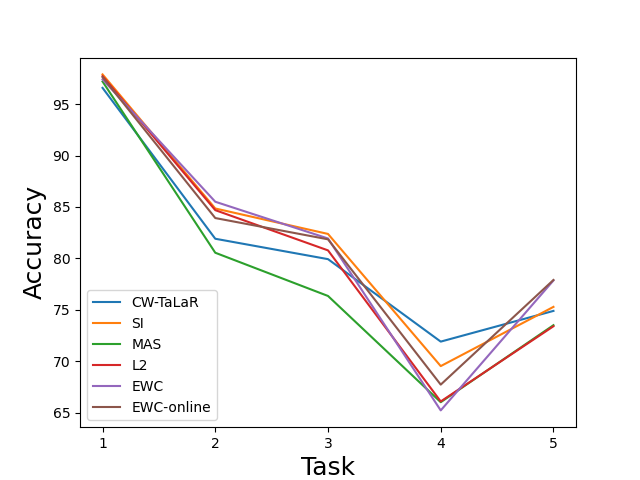}
& \ 
\includegraphics[width=0.33\textwidth]{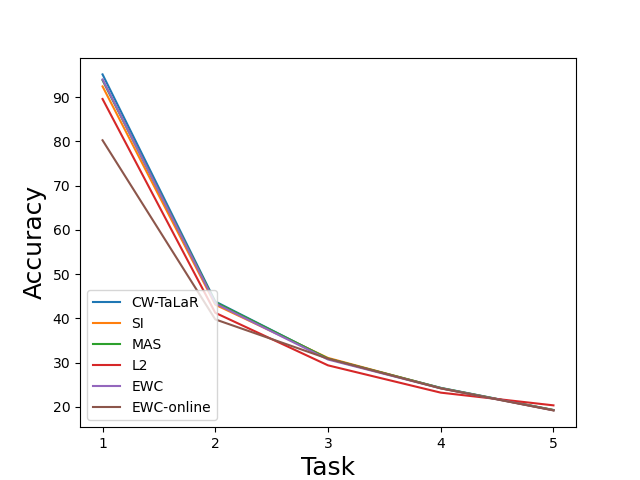}
\end{tabular}
\caption{Comparison of average accuracy over currently completed tasks achieved by each method applied for Split CIFAR-10 in three CL scenarios, calculated on validation dataset during the training process. (All results were additionally averaged over 10 runs.) Note that scores for CW-TaLaR are comparable with those for SI and MAS.}
\label{fig:com_forget_cifar}
\end{figure}

We summarize the results of all experiments performed on the Split CIFAR-10 dataset in Table \ref{split_cifar_table}. Note that \our{} in all scenarios gives scores comparable to SI and MAS, although works slightly worse than (online) EWC. This observation seems to be confirmed by Figure \ref{fig:com_forget_cifar}, which shows comparison of efficiency of all considered CL strategies in overcoming catastrophic forgetting during the training process. We suppose that such loss of dominance when compared to the results for Split MNIST is due to significant change in applied architecture (here we use CNN instead of MPL).

\begin{table}[!t]
\begin{center}
\caption{Average accuracy (\%, higher is better) with standard deviation, achieved on validation dataset by each method applied for Split CIFAR-10 in three CL scenarios, calculated over 5 tasks and 10 runs (at the end of the training process).}
\vskip2mm
\begin{tabular}{cccc}
\toprule
  Method & ITL &  IDL & ICL \\
\midrule
   Adam        & 72.65 $\pm$ 3.39 & 73.13 $\pm$ 1.39 & 19.31 $\pm$ 0.05 \\
   SGD         & 68.93 $\pm$ 5.68 & 73.13 $\pm$ 1.35 & 17.51 $\pm$ 2.42 \\
   Adagrad     & 70.90 $\pm$ 6.50 & 72.57 $\pm$ 1.54 & 16.02 $\pm$ 0.54 \\
   L2          & 76.30 $\pm$ 0.96 & 73.40 $\pm$ 1.71 &\bf 20.36 $\pm$ 2.13 \\ 
\midrule
   EWC         & 86.07 $\pm$ 4.06 &\bf 78.16 $\pm$ 0.68 & 19.24 $\pm$ 0.11 \\
   Online EWC  &\bf 87.57 $\pm$ 3.89 & 77.46 $\pm$ 0.89 & 19.22 $\pm$ 0.12 \\
   SI          & 84.41 $\pm$ 2.60 & 75.28 $\pm$ 1.26 & 19.27 $\pm$ 0.10 \\
   MAS         & 82.13 $\pm$ 1.80 & 73.50 $\pm$ 1.54 & 19.30 $\pm$ 0.04 \\ 
  CW-TaLaR (ours) & 82.09 $\pm$ 2.04 & 74.89 $\pm$ 0.61 & 19.20 $\pm$ 0.08 \\
\midrule
   Offline (upper bound) & 95.25 $\pm$ 0.64 & 91.35 $\pm$ 0.32 & 84.83 $\pm$ 1.60  \\
\end{tabular}
\label{split_cifar_table}
\end{center}
\end{table}

\subsection{Permuted MNIST}

\begin{figure}[!t]
\centering
\begin{tabular}{@{}c@{}c@{}}
 Accuracy &  Average accuracy  \\
 on individual tasks  &   over currently completed tasks  \\
 \includegraphics[width=0.5\textwidth]{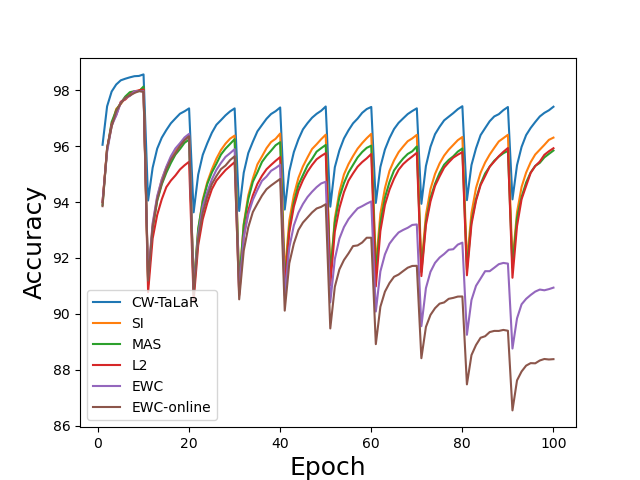}
  & \
  \includegraphics[width=0.5\textwidth]{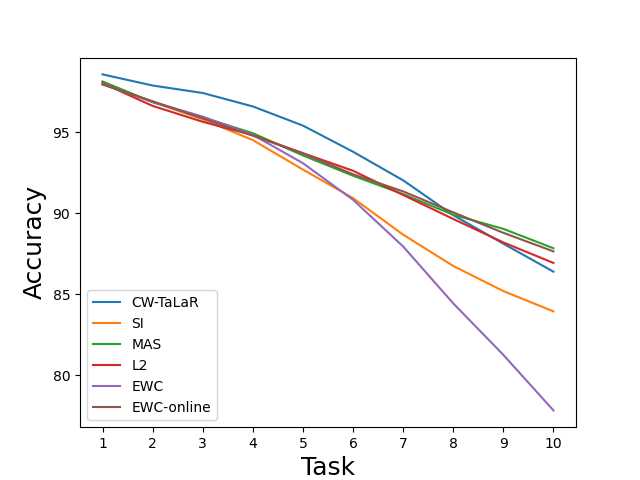}
 \end{tabular}
  \caption{Comparison of average accuracy  over currently completed tasks achieved  by each method applied for Permuted MNIST in IDL scenario, calculated on validation dataset during the training process. (All results were additionally averaged over 10 runs.) Note that CW-TaLaR is superior in learning new tasks or for relatively short term scenarios.}
  \label{fig:permuted_MNIST_forgetting}
\end{figure}

We summarize the results of IDL experiments performed on the Permuted MNIST dataset in Table \ref{permuted_mnist_table}.
Note that almost all obtained scores (excluding those for EWC) are comparable. However, closer inspection of learning process, which is shown in Figure \ref{fig:permuted_MNIST_forgetting}, allows to guess that the \our{} method is superior in learning new tasks, although may slightly suffer from forgetting in the case when the number of tasks is relatively large.

\begin{table}[!t]
\begin{center}
\caption{Average accuracy (\%, higher is better) with standard deviation, achieved on validation dataset by each method applied for Permuted MNIST in IDL scenario, calculated over 5 tasks and 10 runs (at the end of the training process).}
\vskip2mm
\begin{tabular}{cc}
\toprule
  Method &  IDL \\
\midrule
  Adam          & 76.19 $\pm$ 2.17   \\
  SGD           & 70.35 $\pm$ 1.53   \\
  Adagrad       & 85.24 $\pm$ 0.75   \\
  L2            & 86.92 $\pm$ 1.68   \\ 
\midrule
  EWC           & 77.81 $\pm$ 1.32  \\
  Online EWC    & 87.63 $\pm$ 0.99   \\
  SI            & 83.92 $\pm$ 1.22     \\
  MAS           & \bf 87.82 $\pm$ 1.47   \\ 
  CW-TaLaR (ours)  & 86.38 $\pm$ 1.08  \\
\midrule
  Offline (upper bound) & 97.83 $\pm$ 0.05
\end{tabular}
\label{permuted_mnist_table}
\end{center}
\end{table}

\subsection{Architecture}

For all models we performed a grid search over hyperparameters. Our classification networks were trained on 128-element mini-batches using ADAM optimizer (see \cite{kingma2017adam}) with a starting learning rate 1e-4, which decreases to 1e-5 after the first task. CW generator was also trained on 128-element mini-batches using ADAM optimizer with learning rate set to 1e-4. We applied ReLU as the activation function and the loss function (in all methods and all scenarios) is a standard cross-entropy for classification frameworks. Below we include architecture details for networks used in various experimental settings.

\subsubsection{Split MNIST}
Each classification task was trained for 4 epochs. Generator was trained for 10 epochs.
\begin{itemize}
\setlength\itemsep{0.0em}
\item[]\textbf{classifier}
\begin{itemize}
\item[] fully connected networks with ReLU activation functions
\item[] layer sizes: 784 $\to$ 1024 $\to$ 512 $\to$ 256 $\to$ 2/10 neurons depends on setting
\item[] softmax output layer
\end{itemize}
\item[]\textbf{generator} 
\begin{itemize}
\item[] noise dimension: 8
\item[] fully connected networks with ReLU activation functions
\item[] layer sizes: 8 $\to$ 16 $\to$ 32 $\to$ 64 $\to$ 128 $\to$ 256 
\item[] ReLU output layer
\end{itemize}
\end{itemize}

\subsubsection{Permuted MNIST}
Each classification task was trained for 10 epochs. Generator was trained for 10 epochs. Architectures used for generator and classifiers were the same as those described for Split MNIST.

\subsubsection{Split CIFAR-10}

Each classification task was trained for 12 epochs. Generator was trained for 10 epochs. 

\begin{itemize}
\setlength\itemsep{0.0em}
\item[]\textbf{classifier}
\begin{itemize}
\item[] all convolutional layers use $3 \times 3$ kernel and are followed by batch normalization and ReLU activation
\item[] two convolutional layers with $32$ channels 
\item[] convolutional layer with $64$ channels
\item[] max pooling layer
\item[] two convolutional layers with $64$ channels 
\item[] convolutional layer with $128$ channels
\item[] max pooling layer
\item[] two convolutional layers with $128$ channels 
\item[] convolutional layer with $256$ channels
\item[] max pooling layer
\item[] two dense ReLU layers with $256 \times 4 \times4$ and $256$ neurons
\item[] dense layer with 2 or 10 outputs depending on scenario
\item[] softmax output layer
\end{itemize}
\item[]\textbf{generator} 
\begin{itemize}
\item[] noise dimension: 64
\item[] fully connected networks with ReLU activation functions
\item[] layer sizes:  64 $\to$ 128 $\to$ 256 $\to$ 512 $\to$ 256 
\item[] ReLU output layer
\end{itemize}
\end{itemize}

\section{Related work}

As we indicated in Introduction the CL strategies can be assigned into three fuzzy collections: architectural strategies, (pseudo) rehearsal strategies and regularization strategies.

Such methods as Progressive Neural Networks (PNN, see \cite{rusu2016progressive}) or CopyWeights with Re-init (CWR, see \cite{CWR}) prevent catastrophic forgetting by changing their architectures. CWR trains in each task the same model with new parameters and adds them to the old ones. A similar idea was used in PNN, where in the next task a bigger neural network is trained, but a part of its weights is taken from the previous neural network. However, since our method cannot be considered as an architectural strategy, we do not feel obligated to compare it to such CL models.

Full rehearsal method (mixing all older examples with new ones) is rather a theoretical solution. An example of effective rehearsal strategy was proposed by \cite{hayes2019memory}. It creates a special buffer for each label and each buffer aggregates the same number of examples. The old examples make a place for new ones by compression or removal. Up to some extent, CW-TaLaR algorithm can be considered as a pseudo-rehearsal method. However, instead of aggregating old examples, a generative model is trained so we need to only remember all its parameters.

Another approach to solving CL problems (Learning without Forgetting, LwF) was made by \cite{Li2016}.
 LwF relies on controlling a change of output data. It was achieved by regularization but also by  adding a small number of nodes in each task. Therefore, it could be partially considered as an architectural strategy.

A pure regularization strategies usually assume that the optimal parameters for a new task can be found in the neighborhood of those obtained for older tasks. For example, this idea is underlined by Elastic Weight Consolidation (EWC, see \cite{Kirkpatrick3521}). EWC algorithm controls the change of parameters only by introducing regularization term in a loss function. The importance of parameters is weighted by the diagonal of the Fisher information matrix.
An alternative for this method was presented by \cite{Kolouri2020Sliced}.
There the respective diagonal matrix in regularization term originated from the Cramer distance between output densities for old and new tasks.
Another important modification of EWC is Synaptic Intelligence (SI, see \cite{Zenke2017}). The novel of this method lies in detecting the change of parameters in the previous tasks and weighting a regularization term by their importance. 

For the other significant CL models we refer the reader to \cite{ CWR,distillation,GEM,iCaRL, GDM,MISHRA202151,nguyen2018variational,vandeven2019scenarios,8545895,CLsurvey}.

\section{Conclusion}

In this paper, we presented the new CL learning strategy (\our{}) based on regularization involving the Cramer-Wold distance between target layer distributions representing current and past information, which does not suffer from growing memory consumption. We applied our method for a few supervised CL scenarios using various versions of experimental setup proposed in \cite{hsu2018re}.

Our experiments show that \our{} gives superior average accuracy scores (when compared to EWC, SI and MAS methods) for IDL and ICL scenarios on Split MNIST dataset, and very competitive results for the other considered settings (including ITL) on Split MNIST, Split  CIFAR-10 and Permuted MNIST datasets.

In conclusion, the proposed method works very well in general but seems to be especially efficient for CL scenarios with a single-head classifiers
(i.e IDL and ICL). However, the closer inspection of the results for Permuted MNIST and Split CIFAR-10 suggests that it may slightly suffer from forgetting, when either the number of tasks is relatively large or the applied architecture consists of convolutional (rather then fully connected) layers, although in these cases the average scores still remain competitive with respect to the other state-of-the-art CL models.

\section{Acknowledgments}
The research of S. Knop was funded by the Priority Research Area Digiworld under the program Excellence Initiative -- Research University at the Jagiellonian University in Krakow. 
The research of P. Spurek was funded by the Priority Research Area SciMat under the program Excellence Initiative – Research University at the Jagiellonian University in Krakow. 

\bibliographystyle{unsrtnat}
\bibliography{CW-TaLaR_arXiv}

\end{document}